\documentclass[runningheads]{llncs}
\usepackage[T1]{fontenc}
\usepackage{algorithm}
\usepackage{algpseudocode}
\usepackage{amsmath}
\usepackage{graphicx}
\begin{document}
\title{C2F-CHART:
A Curriculum Learning Approach to Chart Classification}

%
%

\author{Nour Shaheen\inst{1,2} \and
Tamer Elsharnouby\inst{2} \and
Marwan Torki\inst{1,2}}

\authorrunning{N. Shaheen et al.}
%
\institute{Faculty of Engineering, Alexandria University, Egypt\and
Applied Innovation Center, MCIT, Egypt \\
\email{nourheshamshaheen@gmail.com, tamer.elsharnouby@aic.gov.eg, mtorki@alexu.edu.eg} 
}
\maketitle              
\begin{abstract}
In scientific research, charts are usually the primary method for visually representing data. However, the accessibility of charts remains a significant concern. In an effort to improve chart understanding pipelines, we focus on optimizing the chart classification component. We leverage curriculum learning, which is inspired by the human learning process. In this paper, we introduce a novel training approach for chart classification that utilizes coarse-to-fine curriculum learning. Our approach, which we name C2F-CHART (for coarse-to-fine) exploits inter-class similarities to create learning tasks of varying difficulty levels. We benchmark our method on the ICPR 2022 CHART-Infographics UB UNITEC PMC dataset, outperforming the state-of-the-art results. 

\end{abstract}
\section{Introduction}

Charts are commonly used to represent features and relationships in data. They are also regularly used in scientific research. In the areas of machine learning, a researcher has to interpret loss curves, confusion matrices, data analysis plots, feature importance plots, and others. However, when dealing with visual representations, there is always the risk that individuals with vision impairment, low vision, or blindness are at a disadvantage. To increase the accessibility of charts, which are inherently visual, automatic pipelines for chart data extraction are needed. The chart data extraction process is called chart mining \cite{9085944}. Often, the first step in this process is high-level chart classification. The division of chart images into specific categories can simplify further processing steps in the pipeline. This initial categorization can allow the following steps to either leverage the chart type, as meta-information about the image, or to assign different processing methods for each type instead of using the same method arbitrarily for all charts. 

Past research \cite{thiyam2023chart,dhote2023survey,amara2017conv} has investigated the use of deep learning methods for image classification and contrasted them to achieve a robust, highly accurate chart classifier. Most of the research has been directed toward identifying the best model architecture for the task, whether a convolutional neural network (CNN) architecture or a transformer architecture.

Inspired by how humans learn, we use a curriculum learning (CL) \cite{10.1145/1553374.1553380} based approach. We extend the coarse-to-fine CL algorithm \cite{stretcu2021coarse}, which focuses mainly on classification tasks. Stretcu et al. \cite{stretcu2021coarse} argue that, during classification, we can attribute the model’s errors to similarities in class labels. Our motivation to use this approach as a building block is because one of the existing challenges in chart classification resides in the similarities between the different chart classes, or the \textit{output space} of the model. 

\begin{figure}
\centering
\includegraphics[width=11cm]{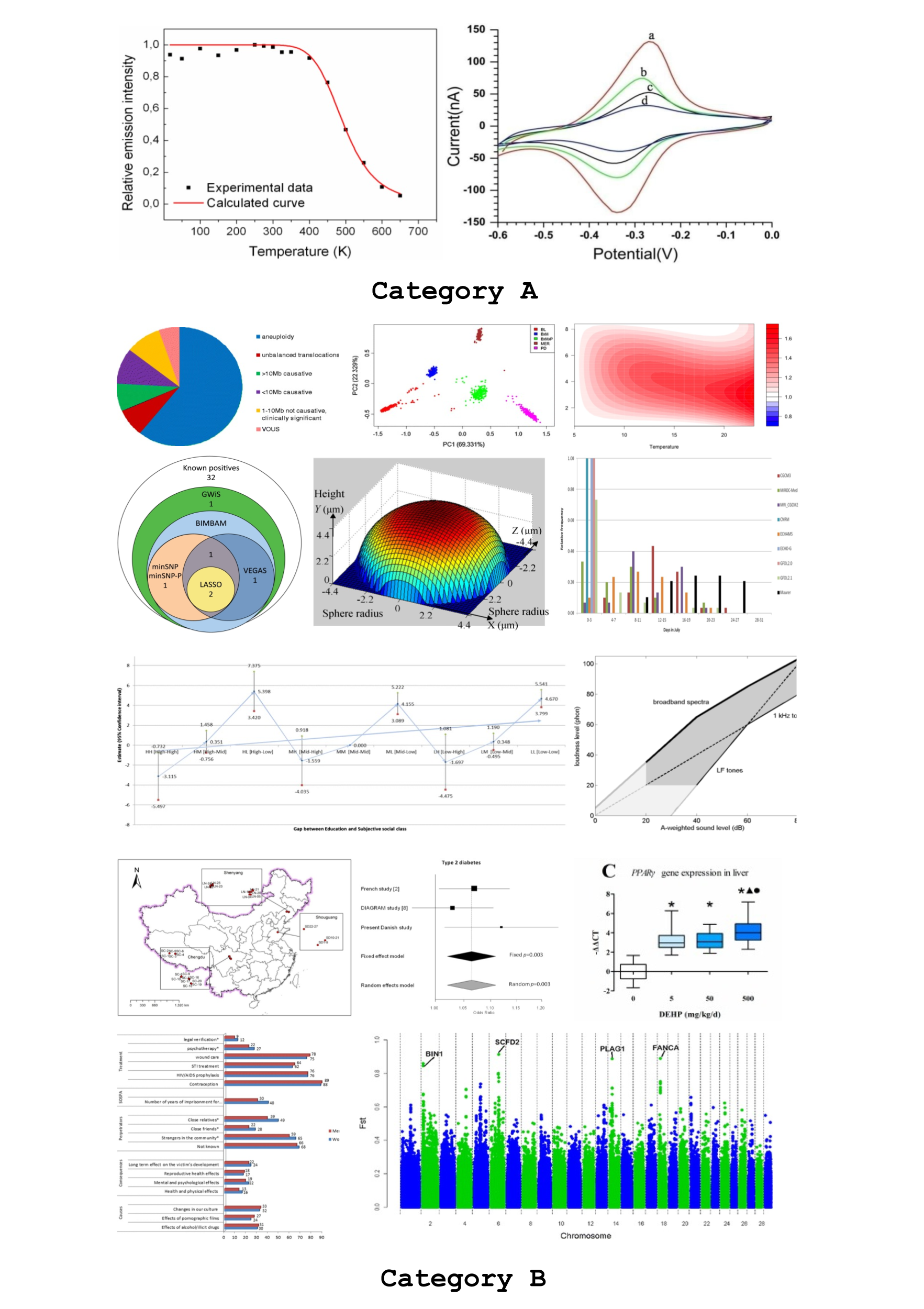}
\caption{Visualization of chart type clustering to construct tasks of varying difficulty.} \label{intro}
\end{figure}

This inter-class similarity, usually considered a challenge, is leveraged by our training approach. Our CL setting allows us to construct learning tasks that are guaranteed to vary in difficulty by grouping similar classes. This allows us to construct simpler tasks, where we classify broader categories, and then to construct more complex ones, where we focus on distinguishing between specific types and classes. We can visualize this in Fig.~\ref{intro}. 
We refine the learning process across multiple levels of complexity and then combine the experience of multiple different learners at each level. We show that this optimizes the model's ability to discern nuanced differences in chart features that each learner might have picked up independently.

The contributions of this paper are as follows: 
\begin{itemize}
    \item We developed a novel training approach that, to the best of our knowledge, has not been used in chart classification before. We then used our approach to train the current state-of-the-art model architecture \textit{Swin-Chart} \cite{dhote2023survey}.
    \item We ran an evaluative analysis to confirm that our approach outperforms the SOTA architecture on the ICPR 2022 CHART-Infographics UB Unitec PMC Dataset \cite{9956289}. It also outperforms the ICPR 2022 CHART-Infographics competition winner on the same dataset. Our analysis also proved that our method exceeded the results of traditional coarse-to-fine CL.
\end{itemize}
The structure of the paper is as follows. In Section 2, we provide a concise overview of the past research conducted in chart classification using deep learning methods as well as curriculum learning techniques for image classification. Section 3 describes the dataset we used for benchmarking our results and why we selected it. In Section 4, we detail the method we developed. Our results, experiments, and comparison with other methods are provided in Section 5. Lastly, Section 6 concludes the paper and presents directions for future study.

\section{Related Work}
\subsection{Deep Learning Methods for Chart Classification}

This section focuses on previous work investigating the use of various deep learning methods for chart classification. We acknowledge that the use of both model-based methods and traditional machine learning methods for chart classification has been thoroughly investigated \cite{huang2003model,amara2017conv,thiyam2023chart}, yet each approach has its challenges.

Unlike a model-based approach, which models only a predefined set of chart types, deep learning methods are agnostic to the specific types encountered. They’re also superior to traditional machine learning methods that require hand-crafted features. Also, these methods aren’t guaranteed to generalize well due to the varieties in chart image datasets. 

Regarding deep learning methods, convolutional neural networks (CNNs) \cite{726791} have long been a staple of image classification in different domains. The ability of CNNs to capture hierarchical features from images through convolutional filters has allowed them to be extremely effective in feature extraction. 

Liu et al. \cite{7333872} introduced DeepChart, which combined CNNs for feature extraction and Deep Belief Networks (DBNs) for classification. Amara et al. \cite{amara2017conv} used a vanilla CNN-based model, inspired by the LeNet architecture \cite{726791} and tested it on their own dataset which is composed of 11 categories. Many subsequent papers \cite{s20164370,thiyam2023chart} compare other CNN architectures such as different layer versions of ResNets \cite{7780459}, DenseNets \cite{huang2018densely}, VGG Networks \cite{simonyan2015deep}, Xception Modules \cite{chollet2017xception}, and EfficientNet \cite{tan2020efficientnet}. 

In 2021, Bajic et al. \cite{jimaging7110220} introduced a new addition to the CNN method of classifying charts: a Siamese CNN. They argue that when small datasets are used, a Siamese CNN outperforms a classic CNN in both classification accuracy and F1-score.

Finally, Dhote et al. \cite{dhote2023survey} compared the use of several CNN architectures for this task on the same testing dataset. They compared and contrasted the ResNet-152, the Xception module, the DenseNet-121, and ConvNeXt \cite{liu2022convnet}, concluding that the Resnet-152 achieved the highest performance on the ICPR 2022 UB UNITEC PMC testing dataset out of all the other CNN architectures. 

However, with the advent of image transformer models \cite{dosovitskiy2021image}, it’s fair to say CNNs have been surpassed in performance. Vision transformers treat images as sequences of patches. Instead of the localized feature maps produced by CNNs, transformer models leverage the self-attention mechanism to capture global dependencies. Dhote et al. \cite{dhote2023survey} compared two transformers backbones: Swin-based and DeIT-based, and concluded that Swin transformers with different patch sizes outperform CNN-based architectures. Their state-of-the-art chart classification model, Swin-Chart, was, to our knowledge and previous to this work, the best performing transformer model architecture for chart classification on the aforementioned dataset.

\subsection{Curriculum Learning}

Curriculum learning (CL) was first introduced by Bengio et al. \cite{10.1145/1553374.1553380}. The intuition for it stemmed from the methods used by humans to learn information. Around the world, humans start by learning easier concepts before gradually moving towards more complex concepts later.  Usually, the input data used to train machine learning models is not organized in any meaningful way. The samples are instead fed to the model in a random order, with easy and difficult samples shuffled and presented to the model with no heed to its training status or the difficulty of each data point.

Soviany et al. \cite{soviany2022curriculum} propose in their survey of CL methods that increasing the complexity of the data, referred to as the \textit{experience} of the model, is not the sole approach to implement curriculum learning. They contend that complicating any other machine learning component will produce a more involved objective function. Namely, this might be done by increasing the complexity of the model itself, by adding or activating neural units for example, or by increasing the complexity of the class of tasks the model is being trained on.

Coarse-to-fine curriculum learning \cite{stretcu2021coarse}, which is the main building block of our method, keeps the experience of the model consistent during training, but instead leverages the similarity between data classes to define a set of tasks $\{f_0, f_1, ...\}$ that are guaranteed to vary in difficulty. 

The inspiration behind this coarse-to-fine technique originated from a specific aspect of human learning, where humans learn to break down specific, detailed tasks into simple milestones. Stretcu et al. \cite{stretcu2021coarse} illustrate this method with the analogy of a child initially learning to identify dogs broadly as dogs, before later learning to differentiate between different dog breeds. Instead of relying on varying difficulty levels in the input data, the method itself ensures variability by progressively introducing tasks of increasing complexity as the model continues learning. This is achieved by clustering similar classes into broader categories, creating a hierarchical structure of class labels. Each task assigned to the model corresponds to a level within the hierarchy, with simpler tasks being at the top, where the categories are less specific. 

\section{Dataset}
For training and testing, we used the datasets provided for the ICPR 2022 CHART-Infographics competition \cite{9956289}. Both the training and testing sets are comprised of real charts from the PubMed Central that have been manually annotated. The testing set for the chart classification task, called \textit{Split 1} in the data and the competition paper, is composed of 11,388 samples while the training set is composed of 22,923 samples. Table~\ref{data} describes the frequency of each class in both sets. We reserved 10\% of the dataset's training split for validation purposes and only used 90\% throughout our training process.

\begin{table}
\centering
\caption{Frequency of each chart type in the ICPR 2022 UB PMC Dataset}\label{data}
\begin{tabular}{|l|c|c|}
\hline
Chart Type &  Train & Test \\
\hline
Area &  172 & 136 \\
Bar (horizontal) &  787 & 425\\
Bar (vertical) & 5,454 & 3,183\\
Box (vertical) & 763 & 596\\
Heatmap & 197 & 180\\
Interval (horizontal) & 156 & 430\\
Interval (vertical) & 489 & 182\\
Line & 10,556  & 2,776\\
Manhattan & 176 & 80\\
Map & 533 & 373\\
Pie & 242 & 191\\
Scatter & 1,350 & 949\\
Scatter-line & 1,818 & 1,628\\
Surface & 155 & 128\\
Venn & 75 & 131\\
\hline
Total & 22,923 & 11,388 \\
\hline
\end{tabular}
\end{table}

Another important motivator for selecting this dataset was the desire to conduct a comparative analysis with other research and quantify the improvements made. Since this dataset was used in the CHART-Info competition, we have real results of different deep learning methods to compare against.

\section{Method}

\begin{figure}
\centering
\includegraphics[width=9cm]{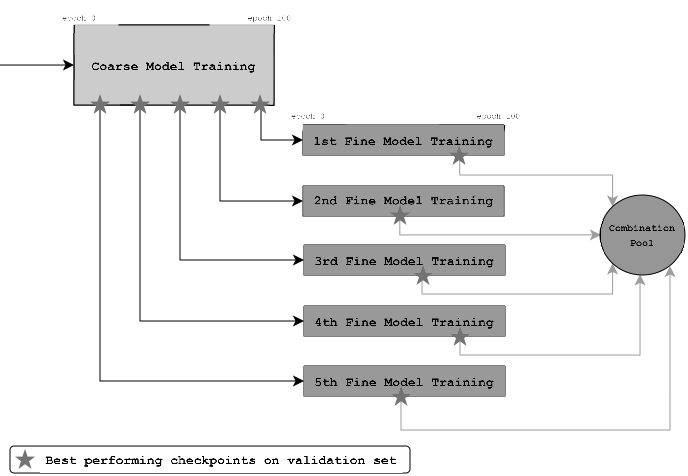}
\caption{Overview of our curriculum learning method.} \label{fig2}
\end{figure}

In this section, we present our hierarchical coarse-to-fine CL approach, leveraging a Swin Transformer model for chart classification. We previously described how traditional CL and coarse-to-fine CL differ in how they consider difficulty. Coarse-to-fine CL's main idea is to create tasks with increasing difficulty $\{f_0, f_1, ...\}$ while not changing the order of the input data. However, our approach, which is described by Fig.~\ref{fig2}, goes a step further and considers at which point in the learning process we should start teaching the model the more complex task. We also consider knowledge sharing between learners who shifted to the complex task at different points in that process. When the model is learning the simpler task $f_0$, the traditional approach would have us transferring knowledge to the more complex task $f_1$ after a certain time (number of epochs) or when it performs best on $f_0$ (checkpoint with the highest validation score). Instead, we argue that choosing either the best performing model on $f_0$ or the final model after several epochs does not necessarily produce optimal results. 

Instead, we choose to transfer knowledge from the top-$K$ learners of $f_0$ so that each learner is then trained on $f_1$, producing $K$ training paths. We then choose the top learner of each path of $f_1$, producing $K$ final checkpoints. We argue that the subsequent sharing of the knowledge obtained by these $K$ final checkpoints produces better results. Knowledge sharing here happens during inference time.  

Our method involves three main steps: clustering, training/fine-tuning, and ensembling. We use minimizing cross-entropy (CE) loss as an objective function. We provide the detailed pseudocode of these three steps in Algorithm~\ref{alg:coarse_to_fine}.

\begin{algorithm}
\caption{Hierarchical Coarse-to-Fine CL with Swin-Chart}
\label{alg:coarse_to_fine}
\begin{algorithmic}
\State \textbf{Input:} Swin Transformer $\theta$ with input image dimension 224 (SwinL\_224) 
\State Compute class hierarchy $H$ using an auxiliary clustering function

\State Define auxiliary objective functions $f_0$ and $f_1$:
\State \quad $f_0$: Minimize CE loss for the 2 classes in level 1 of $H$
\State \quad $f_1$: Minimize CE loss for the 15 classes in level 2 of $H$

\State Initialize $\theta^0$ with SwinL\_224 weights pre-trained on ImageNet
\For{epoch $= 1$ to $100$}
    \State Train $\theta^0$ on $f_0$
    \State Validate on holdout set
\EndFor
\State Select top 5 checkpoints $\{\theta_0^0, \theta_1^0, \theta_2^0, \theta_3^0, \theta_4^0\}$ based on val. F1-scores

\For{each $\theta_i^0$}
    \State Initialize $\theta^1_i$ with encoder parameters from $\theta_i^0$ and random decoder parameters
    \For{epoch $= 1$ to $100$}
        \State Train $\theta^1_i$ on $f_1$
        \State Validate on holdout set
    \EndFor
\EndFor
\State Add top checkpoint from each $\theta^1_i$ to final model pool to combine $\{\theta_0^1, \theta_1^1, \theta_2^1, \theta_3^1, \theta_4^1\}$
\State Conduct combinatorial search to find optimal model combination
\State \textbf{Output:} Model combination with max. F1-score on holdout set
\end{algorithmic}
\end{algorithm}

\begin{figure}
\centering
\includegraphics[width=8cm]{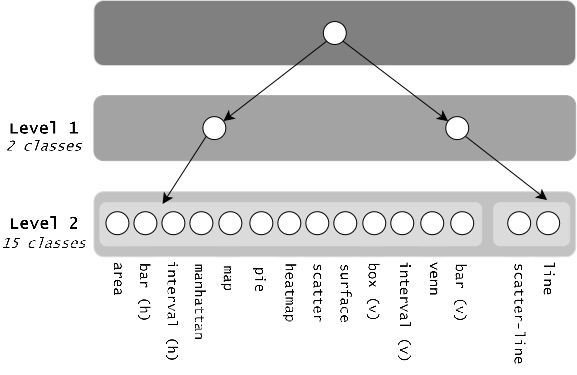}
\caption{Automatically computed coarse-to-fine class hierarchy of 2 levels on the 15 classes of the ICPR 2022 CHART-Infographics UB Unitec PMC Dataset.} \label{fig1}
\end{figure}

\subsubsection*{Step 1: Cluster}
To obtain a hierarchy of simple-to-complex tasks, we needed to first cluster classes based on similarity. We started by training a vanilla classification model using the current state-of-the-art architecture which, as described in \cite{dhote2023survey}, is a Swin Transformer pre-trained on the ImageNet dataset with an input size of 224 that they call Swin-Chart. We use this model to compute a coarse-to-fine class hierarchy $H$ shown in Fig.~\ref{fig1}. $H$ depends on the similarity between the columns of the projection matrix in the output layer or \textit{the predictor} of the Swin model. It is computed through affinity clustering, using the pairwise cosine distances between the columns as a distance matrix \cite{stretcu2021coarse}. The pseudocode for computing $H$ and generating hierarchical clusters is provided in Algorithm~\ref{alg:generate_hierarchy}.

\begin{algorithm}
\caption{Generate Clusters Per Level for Hierarchical Coarse-to-Fine CL}
\label{alg:generate_hierarchy}
\begin{algorithmic}
\State \textbf{Input:} Number of classes $K$, training data, pre-trained baseline Swin model $\theta$
\State Train Swin model $\theta$ on the training data
\State Extract the class embeddings from the final layer of $\theta$
\State Let $W \in {R}^{E \times K}$ be the weight matrix of the final layer
\State Compute the distance matrix $D$ using cosine distances between class embeddings:
\For{$k_1 = 1$ to $K$}
    \For{$k_2 = k_1 + 1$ to $K$}
        \State Compute cosine distance $d(k_1, k_2) = 1 - \cos(W_{\cdot k_1}, W_{\cdot k_2})$
        \State Update $D[k_1, k_2] \leftarrow d(k_1, k_2)$, $D[k_2, k_1] \leftarrow d(k_1, k_2)$
    \EndFor
\EndFor
\State Apply affinity clustering on the distance matrix $D$ to form the hierarchy $H$
\State Initialize $\text{clustersPerLevel} \leftarrow []$
\For{$l = 1$ to $\text{depth}(H)$}
    \State Initialize $\text{clustersPerLevel}[l] \leftarrow []$
    \For{each node $n \in H.\text{nodesAtDepth}(l)$}
        \State Create cluster $c$ by grouping leaves of the sub-tree rooted at $n$
        \State Append $c$ to $\text{clustersPerLevel}[l]$
    \EndFor
\EndFor

\State \textbf{Output:} Class hierarchy $H$, clusters per level $\text{clustersPerLevel}$
\end{algorithmic}
\end{algorithm}

\subsubsection*{Step 2: Train, Divide, and Fine-tune}
Using our two-level class hierarchy, we constructed two auxiliary tasks $f_0$ and $f_1$, one for each level. We define the objective function $L_i$ used to train the $i$-th auxiliary task $f_i$ as optimizing the maximum likelihood by minimizing cross-entropy (CE) loss, which is equivalent to minimizing the negative log-likelihood: 
\begin{equation}
L_i = - \sum_{j} \log \left( \sum_{c \in C_i(y_j)} \exp \{ f_{i}(x_j) \} \right)
\end{equation}
where $i$ denotes the hierarchy level we're working at. For each sample $j$, $x_j$ and $y_j$ represent the input data and its corresponding true label. The term $C_i(y_j)$ denotes the cluster in level $i$ that the class $y_j$ belongs to. 

We started by initializing a new instance of the Swin-Chart architecture and training it on $f_0$ to produce top-$K$ checkpoints from different points of the training journey, judged based on the average per-class F1-measure of a hold-out validation set. We chose K = 5, and so we obtained 5 level-1 models [$\theta_0^0$ - $\theta_4^0$] in order of validation scores. The output space of each $\theta_i^0$ is simply the two main clusters in level 1, as shown in Fig.~\ref{fig1}.

For each $\theta_i^0$, we used the staged coarse-to-fine CL algorithm \cite{stretcu2021coarse}: we initialized five Swin-Chart level-2 models [$\theta_0^1$ - $\theta_4^1$], whose encoder parameters were set as the encoder parameters of the corresponding $\theta_i^0$ and whose predictor parameters were randomly initialized. We then fine-tuned all parameters of our level-2 models on the desired output space, the 15 classes in level 2, as shown in  Fig.~\ref{fig1}.

After the second fine-tuning step, we chose the checkpoint with the maximum validation score from each level-2 model as an ingredient for the final combination step, totaling $K$ final models, judged based on the average per-class F1-measure. 

\subsubsection*{Step 3: Combine}
To combine our $K$ ingredients, we considered two approaches: an ensembling method through averaging of predictions, and the model soups method \cite{wortsman2022model}. 
For each method, we conducted a combinatorial search to choose the optimal model combination to ensemble or soup. We judged all combinations in both methods on their validation scores and chose our final model to be the model combination producing the maximum F1-score on the hold-out validation set. 

\section{Experiments}

\subsection{Setup}

As mentioned before, we used the datasets provided for the ICPR 2022 CHART-Infographics competition for both training and testing. We benchmarked our results on the testing dataset, called ICPR 2022 UB Unitec PMC Dataset, and compared them with previous work. Throughout the following experiments, we designated 10\% of the dataset as a hold-out validation set, and used the remaining 90\% in our clustering and fine-tuning steps. 

In order to avoid attributing our improvements to a change in hyperparameters,  we used an identical experimental setup as the one mentioned in \cite{dhote2023survey}. This choice guarantees consistency and allows for a fair comparison between our approach and theirs.

To obtain the class hierarchy $H$ in Fig.~\ref{fig1}, we trained a Swin transformer, pre-trained on ImageNet, for 100 epochs, on a Tesla V100-SXM2-32GB GPU accelerator. We used the Pytorch framework with a learning rate of $1e-4$ and a batch size of 16. Our loss function was label-smoothing cross entropy loss. We used the Adagrad optimizer. After training, we leveraged the columns in the projection matrix at the final layer of our classifier. Using each column's weights as a representation for the corresponding class, we computed the pairwise cosine distances matrix between all classes $D$. We subsequently used $D$ as a similarity measure for affinity clustering to generate $H$.  

We then trained the Swin classifier from ImageNet weights on level-1, whose output space is composed of two classes, with the same settings, for 100 epochs. We chose the parameter $K$ specific to our method as 5, and thus saved the top-5 checkpoints resulting from the level-1 training. For each checkpoint, we initialized a model with the same encoder parameters and with randomized predictor parameters to be trained on level-2, whose output space is composed of 15 classes. We fine-tuned these 5 models as well for 100 epochs, using the same learning rate and batch size, and saved the top achieving checkpoint of each level-2 model.

In the final combination step, we compared the use of model soups and model ensembling. We conducted a combinatorial search to determine the subset of models that yielded the best performance on our validation set. The results of this investigation are mentioned subsequently in our ablation analysis. 

\subsection{Comparative Evaluation}
We proceed to test the best-performing model, which we name C2F-CHART and evaluate our results in comparison with Swin-Chart, the current state-of-the-art method in \cite{dhote2023survey}, other deep learning methods evaluated in \cite{dhote2023survey}, as well as the ICPR 2022 CHART-Inforgraphics competition's results in \cite{9956289}. As shown in table~\ref{finalresults}, our testing precision, recall, and F1-score demonstrate superior performance to all competition participants and Swin-Chart.

\begin{table}
\centering
\caption{Comparative results on ICPR 2022 UB Unitec PMC Dataset. }\label{finalresults}
\begin{tabular}{|l|c|c|c|}
\hline
Team/Method& Recall & Precision & F1-score \\
\hline
Swin-Chart \cite{dhote2023survey} &\textbf{93.3}\%& 93.7\%& 93.2\%\\
IIT\_CVIT \cite{9956289} & 90.1\%& 92.6\%& 91.0\%\\
Resnet-152 \cite{7780459} & 89.9\% & 90.5\% & 89.7\%\\
ConvNeXt \cite{liu2022convnet} & 89.8\% & 90.6\% & 89.6\%\\
UB-ChartAnalysis \cite{9956289} & 88.1\% & 90.0\% & 88.6\%\\
DenseNet-121 \cite{huang2018densely} & 87.9\% & 88.7\% & 87.5\%\\
six\_seven\_four \cite{9956289} & 80.8\% & 86.5\% & 82.7\%\\
CLST-IITG \cite{9956289} & 65.7\% & 70.4\% & 65.4\%\\
\textbf{C2F-CHART (Ours)} &93.17\%& \textbf{95.19\%}& \textbf{93.98}\%\\
\hline
\end{tabular}
\end{table}

\subsection{Ablation Analysis}

To explain our ablation analysis, we showcase the results of level-1 and level-2 model training. 
Table~\ref{levelscore} shows the performance of the top 5 level-1 checkpoints on our validation set. It also shows the top achieving level-2 checkpoint trained from the corresponding level-1 model.  We can observe that our highest validation score in level-2 doesn't necessarily result from the "top achieving" checkpoint at level-1. When we are referring to "top achieving" or "best performing" here, we are indicating the model with the highest F1-score, as evaluated on the hold-out validation set.

\begin{table}
\centering
\caption{Validation F1-scores of the top-5 Level-1 checkpoints and the max. validation F1-scores of the Level-2 model trained from each checkpoint.}\label{levelscore}
\begin{tabular}{|c|c|}
\hline
Top-5 L1 Checkpoints&  Max. Score at L2\\
\hline
\textbf{98.7264\%}& 95.4865\%\\
98.6381\%&  \textbf{96.1122\%}\\
98.5947\%&95.4167\%\\
98.5943\%& 95.6498\%\\
98.5941\%& 95.5715\%\\
\hline
\end{tabular}
\end{table}

Consequently, we define three settings of coarse-to-fine curriculum learning. Setting A represents the traditional curriculum learning approach of fine-tuning our top-achieving level-1 checkpoint, as shown in Table~\ref{levelscore} and then testing its corresponding top-achieving level-2 checkpoint.  Setting B describes taking the top-5 checkpoints trained on level-1, fine-tuning all 5 of them and then testing the top-achieving checkpoint out of all the subsequent models, even if it doesn't result from the level-1 model with the highest score, as is our case. Finally, Setting C describes our method of combination after fine-tuning level-2 using model ensembling on a subset of the 5 final models.

Table~\ref{currtable} compares the three different settings and shows how Setting C achieves the highest F1-score on our validation dataset, as well as the highest precision, recall, and F1-Score on our testing dataset.

\begin{table}
\centering
\caption{Comparison between different curriculum learning settings. The left two columns are on the validation set. The three right columns are on the test set.}\label{currtable}
\begin{tabular}{|l|c|c||c|c|c|}
\hline
Method&  L1 Val. F1-Score& L2 Val. F1-Score& Recall&  Precision&F1-Score\\
\hline
Setting A &  \textbf{98.72\%}& 95.48\%& 92.98\% & 94.67\% &93.53\%\\
Setting B &  98.63\%& 96.11\%& 92.56\% & 94.95\%  &93.6\%\\
\textbf{Setting C 
 (Ours)}& N/A& \textbf{96.27\%}&\textbf{93.17\%}&\textbf{95.19\%}&\textbf{93.98\%}\\
\hline
\end{tabular}
\end{table}

This leads us to consider the optimal approach for combining the models, and for choosing the most suitable subset of models to use, referred to subsequently as \textit{ingredients}. We investigated the use of model soups and simple model averaging for our particular use case. In model soups, we aggregate the weights of the ingredients prior to inference, while in simple ensembling, we average the logits produced by each model in our ingredients pool.

To determine the optimal subset of models for the combination step, we ran a comprehensive combinatorial search on our 5 level-2 models shown in the second column of table~\ref{levelscore}. For every combination of models (2-, 3-, 4-, and 5-model combinations), we calculated the validation F1-score of both model souping and simple ensembling. We also ran an "iterative greedy" version of model soups, as described in \cite{unknown}, where we allowed each ingredient to be added more than once. Finally, we select the combination that achieves the highest validation score as our final model. 
\begin{table}
\centering
\caption{Max. validation F1-scores for each combination of models (using ensembling and souping). Subset is chosen from the 5 models with validation F1-scores: 95.48\%, 96.11\%, 95.41\%, 95.64\%, 95.57\%. }\label{ablation}
\begin{tabular}{|l|c|c|}
\hline
Team/Method& Souping& Ensembling\\
\hline
2-model&95.80\%& 96.00\%\\
3-model& 95.25\%& \textbf{96.28\%}\\
4-model& 95.08\%& 93.13\%\\
5-model& 93.86\%& 95.98\%\\
Iterative greedy&96.11\%& N/A\\
\hline
\end{tabular}
\end{table}

Table~\ref{ablation} compares between the maximum validation F1-scores for each number of models in both the ensembling and souping techniques, along with the validation score obtained through the iterative greedy souping method.  We can conclude that in all combinations, model souping does not outperform the validation score of our highest participating ingredient, while ensembling often does.

\begin{figure}
\centering
\includegraphics[width=8cm]{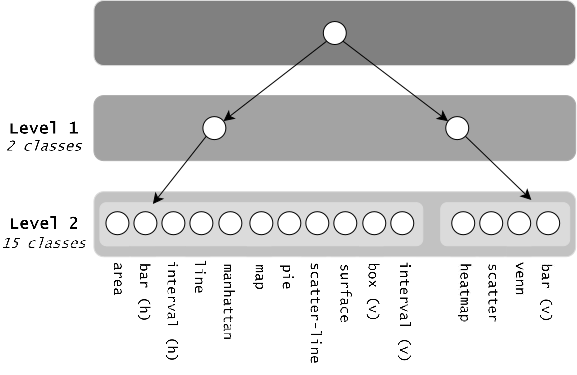}
\caption{Automatically computed coarse-to-fine class hierarchy using a confusion matrix as a distance matrix for clustering.} \label{cluster2}
\end{figure}

Additionally, we investigated the use of another clustering technique to obtain a different hierarchical structure $H$, that we show in Algorithm~\ref{alg:generate_hierarchy_confusion_matrix}. Stretcu et. al. \cite{stretcu2021coarse} also suggested using the confusion matrix of a trained classification model to calculate a distance matrix for the affinity clustering algorithm. Using our vanilla Swin classifier model, we obtained $H$ by estimating the confusion matrix $C$ from our dataset using a hold-out validation set. This involved calculating how often our vanilla model incorrectly predicted each class, and identifying the alternate class predicted instead. Given that $C$ may not be symmetric, we followed the approach outlined in \cite{stretcu2021coarse} and considered our symmetric confusion matrix to be the sum of $C$ and its transpose. We then computed our hierarchy through affinity clustering, using the symmetric confusion matrix as a measure of similarity between classes. This resulted in a different cluster of classes, shown in Fig.~\ref{cluster2}.

\begin{algorithm}
\caption{Generate Class Hierarchy using Confusion Matrix}
\label{alg:generate_hierarchy_confusion_matrix}
\begin{algorithmic}
\State \textbf{Input:} Number of classes $K$, training data, baseline Swin model $\theta$, validation dataset $\{(x_i, y_i)\}_{i=1}^M$
\State Train Swin model $\theta$ on the training data
\State Initialize confusion matrix $C$ of size $K \times K$ with zeros
\For{each $(x_i, y_i)$ in the validation dataset}
    \State Predict class probabilities $\hat{p}(y_i = c \mid x_i; \theta)$ using $\theta$
    \State Increment $C[y_i, c]$ by $\hat{p}(y_i = c \mid x_i; \theta)$
\EndFor
\State Normalize rows of $C$: $C[i, \cdot] \leftarrow C[i, \cdot] / \sum_{j=1}^{K} C[i, j]$
\State Compute symmetric confusion matrix $\hat{C} = C + C^\top$
\State Apply affinity clustering on the symmetric confusion matrix $\hat{C}$ to form hierarchy $H$

\State \textbf{Output:} Class hierarchy $H$
\end{algorithmic}
\end{algorithm}

We re-ran all of our previous experiments on this other cluster and achieved comparable results, shown in tables~\ref{newlevelscore}, \ref{newcurrtable} and \ref{newablation}.

\begin{table}
\centering
\caption{Validation F1-scores of the top-5 Level-1 checkpoints and the max. validation F1-scores of the Level-2 model trained from each checkpoint.}\label{newlevelscore}
\begin{tabular}{|c|c|}
\hline
Top-5 L1 Checkpoints&  Max. Score at L2\\
\hline
\textbf{98.51\%}& 95.96\%\\
98.36\%&  96.02\%\\
98.35\%&95.95\%\\
98.30\%& \textbf{96.32\%}\\
98.25\%& 95.90\%\\
\hline
\end{tabular}
\end{table}

\begin{table}
\centering
\caption{Comparison between different CL settings. The left two columns are on the validation set. The three right columns are on the test set. }\label{newcurrtable}
\begin{tabular}{|l|c|c||c|c|c|}
\hline
Method&  L1 Val. F1-Score& L2 Val. F1-Score& Recall&  Precision&F1-Score\\
\hline
Setting A &  \textbf{98.51\%}& 95.96\%& 92.69\% & 94.03\% &93.14\%\\
Setting B &  98.30\%& 96.32\%& 93.15\%&  94.24\%&93.53\%\\
\textbf{Setting C 
 (Ours)}& N/A& \textbf{96.49\%}&\textbf{93.40\%}&\textbf{94.83\%}&\textbf{93.93\%}\\
\hline
\end{tabular}
\end{table}

\begin{table}
\centering
\caption{Max. validation F1-scores for each combination of models (using ensembling and souping). Subset is chosen from the 5 models with validation F1-scores: 95.96\%, 96.02\%, 95.95\%, 96.32\%, 95.90\%. }\label{newablation}
\begin{tabular}{|l|c|c|}
\hline
Team/Method& Souping& Ensembling\\
\hline
2-model&95.94\%& 96.17\%\\
3-model& 95.99\%& 96.39\%\\
4-model& 95.70\%& \textbf{96.49\%}\\
5-model& 95.13\%& 96.29\%\\
Iterative greedy&96.32\%& N/A\\
\hline
\end{tabular}
\end{table}

\subsection{Qualitative Results}

\begin{figure}
\centering
\includegraphics[width=9cm]{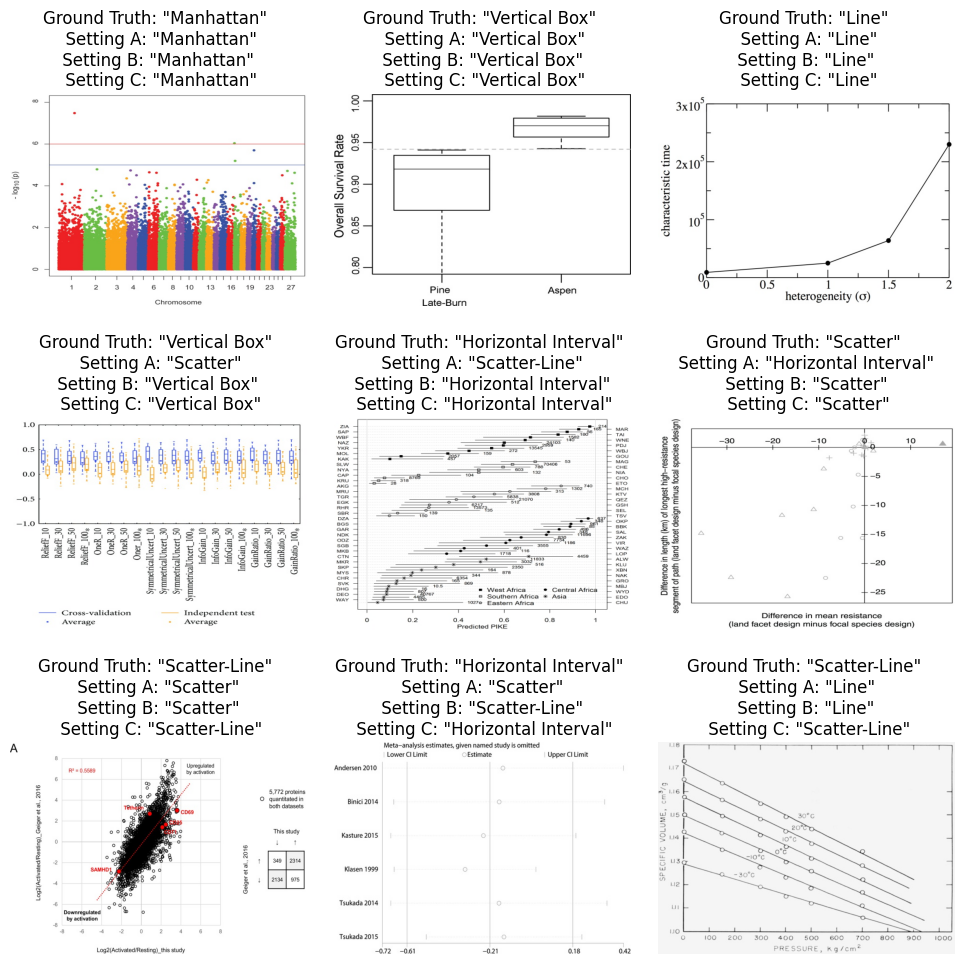}
\caption{Qualitative results comparing Settings A, B \& C.} \label{qual}
\end{figure}

Fig.~\ref{qual} presents qualitative results comparing the three distinct coarse-to-fine CL settings we mentioned previously. The first row showcases the success cases of Setting A, that are also success cases in Settings B and C. In the second row, we showcase selected samples where Setting B exhibits superior performance compared to Setting A. Finally, in the third row, some samples where Setting C surpasses both are displayed.

We can observe from the figure that the charts where Setting C outperforms are charts whose types closely resemble other types. In the first figure of the third row, a short red curve between the thick black scatter dots transforms the figure from a \textit{scatter} plot to the \textit{scatter-line} plot correctly identified by Setting C. In the second chart, the notations identifying horizontal intervals are spaced, and thus were easily confused by Settings A and B as being scatter plot symbols. As well, in the final chart, the scatter plot symbols were misidentified as just indicators on the line, while Setting C correctly identified the figure as a \textit{scatter-line} plot.

\section{Conclusion and Future Work}

We have implemented a novel approach to chart classification using a modified coarse-to-fine curriculum learning algorithm. Our method outperforms the current SOTA approaches on the ICPR 2022 CHART-Infographics UB Unitec PMC Dataset. We compared our method to traditional coarse-to-fine CL, transformer-based, and CNN-based chart classification approaches. Moving forward, we plan to explore the applicability and adaptability of our method beyond the current benchmark, across other datasets with more diverse chart types, to adequately evaluate its usability in real-world scenarios. Also, since our main interest lies in enhancing accessibility for people with visual impairments, we would like to contribute to an end-to-end chart understanding pipeline, which entails extending our research beyond just chart classification to more extensive accessibility features tailored specifically for visually impaired users. 

\section{Acknowledgments}
The authors would like to thank the Applied innovation Center (AIC) of the Egyptian Ministry of Communication and Information Technology for funding the research presented in this paper.

\nopagebreak \bibliographystyle{splncs04}

\bibliography{paper}
\end{document}